\documentclass{article}
\usepackage{spconf,amsmath,graphicx}
\usepackage{enumitem}
\usepackage{bbding}

\usepackage[dvipsnames]{xcolor}
\usepackage{subfigure}
\usepackage{floatrow}
\usepackage{caption}
\usepackage{tikz}
\usepackage{bm}
\usepackage{longtable, array, booktabs}
\usepackage[colorlinks,linkcolor=black]{hyperref}

\title{Towards Using Clothes Style Transfer for Scenario-aware \\ Person Video Generation}
\name{
Jingning Xu$^{1,2}$, Benlai Tang$^2$, Mingjie Wang$^3$, Siyuan Bian$^{2,4}$, Wenyi Guo$^2$, Xiang Yin$^{2}$, Zejun Ma$^2$,
}
\address{
\normalsize  $^1$School of Software Engineering, Tongji University, Shanghai, China \\
\normalsize  $^2$ByteDance AI Lab \\
\normalsize  $^3$School of Computer Science, University of Guelph, ON, Canada \\
\normalsize  $^4$Department of Computer Science and Engineering, Shanghai Jiao Tong University, China
}
\begin{document}
%
\maketitle
\begin{abstract}
Clothes style transfer for person video generation is a challenging task, due to drastic variations of intra-person appearance and video scenarios. To tackle this problem, most recent AdaIN-based architectures are proposed to extract clothes and scenario features for generation. However, these approaches suffer from being short of fine-grained details and are prone to distort the origin person. To further improve the generation performance, we propose a novel framework with disentangled multi-branch encoders and a shared decoder. Moreover, to pursue the strong video spatio-temporal consistency, an inner-frame discriminator is delicately designed with input being cross-frame difference. Besides, the proposed framework possesses the property of scenario adaptation. Extensive experiments on the TEDXPeople benchmark demonstrate the superiority of our method over state-of-the-art approaches in terms of image quality and video coherence. 
\end{abstract}
\vspace{-0.3cm}
\section{Introduction}
\vspace{-0.2cm}
\label{sec:intro}

Inspired by the huge success of Deep Neural Networks, during this decade, the realm of \emph{Clothes Style Transfer} has attracted intensive attention from the research community, which aims to generate realistic person scenes satisfying the demands of diverse attributes~\cite{2020MUST, 2019VTNFP, 2018VITON}. Due to alluring application prospects in controllable person manipulation, virtual try-on clothes texture editing, etc., lots of efforts have been made to improve its performance \cite{2019VTNFP, 2018VITON, 2020Deep, 2020Controllable}. More importantly, video format is a finer carrier for delivering visual experience than static images in many tasks (e.g. person pose animation \cite{2020Deep}). However, the video-based person generation has not been investigated in depth yet by far \cite{2019DwNet}.

Existing approaches for clothes style transfer can be roughly divided into two classes: warping-based \cite{2019VTNFP, 2018VITON, 2020Towards, 2018Human, 2017ClothCap, 2016Detailed} and image generation-based models \cite{2020MUST, 2020Deep, 2020Controllable}. The former type of methods benefit from preserving informative and detailed features of the clothes. Zhang et al. \cite{2018VITON} attempt to estimate a thin plate spline (TPS) transformation for warping the clothing items and refine the coarse results by using a composition mask. Meanwhile, and Yu et al. \cite{2019VTNFP} devise a body segmentation map prediction module to delineate body parts and warped clothing regions. Despite the constant progress achieved by them, it is still  difficult to directly transform the spatially misaligned body parts into the desired shapes due to the non-rigid nature of human body. Hence, these approaches are unable to attain satisfactory performance. To mitigate this issue, several approaches \cite{2020Towards, 2018Human, 2017ClothCap, 2016Detailed} probe into the paradigm of transferring clothes from one person to another by estimating a complicated 3D human mesh and warping the textures to cater for the body topology. Nevertheless, these approaches fail to capture the sophisticated interplay of the intrinsic shape and appearance, and thus results in unrealistic synthesis with deformed textures.

\begin{figure*}[!t]

\centering
\setlength{\abovecaptionskip}{-0.5cm}
\includegraphics[width=16cm]{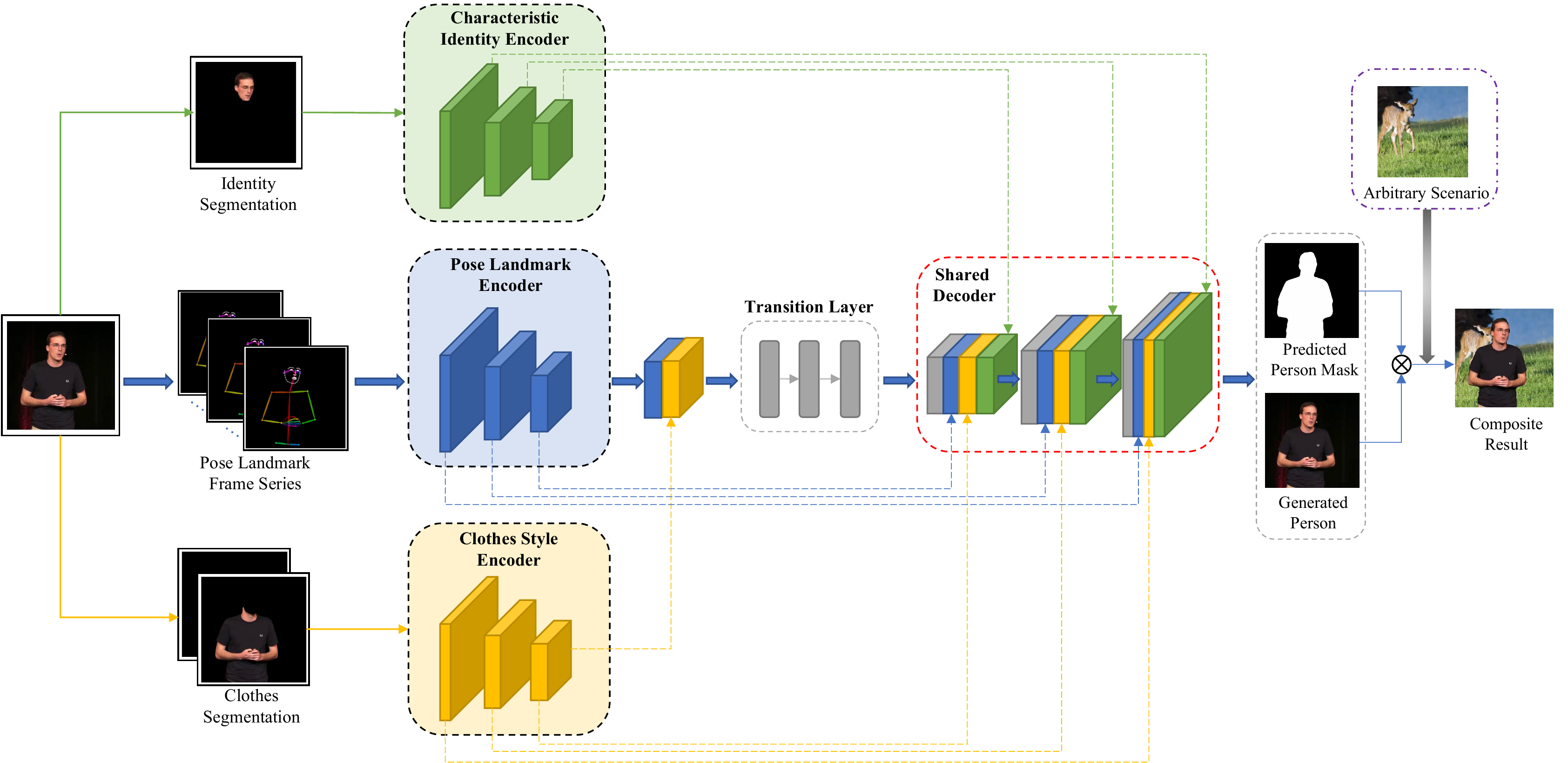}
\caption{The overview of the proposed framework towards using clothes style transfer for scenario-aware video person generation, which is composed of three disentangled encoders (including Pose Landmark Encoder, Characteristic Identity Encoder and Clothes Style Encoder), and a cascaded Shared Decoder. First, these three-path encoders collaboratively represent diverse sources disentangled from the original raw input. Then the separable features are aggregated and fed into the high-level decoder to generate the final output.}
\label{generator}
\end{figure*}

Profiting from the invention of GANs \cite{2016Image, 2017High, 2019Few}, image generation-based approaches have prevailed in image synthesis task, such as  \cite{2020MUST, 2020Deep, 2020Controllable, 2018Everybody, 2017Deformable, 2018A, 2019Progressive}. A core driving force behind these methods is to impose the pose guidance on the image generation process. Hence, the photo-realistic people images with arbitrary poses are rendered, whereas the clothes styles are controlled simultaneously by utilizing various valuable human attributes. Early models~\cite{2018Everybody, 2017Deformable, 2018A, 2019Progressive} only focus on keeping the posture and identity  without providing user manipulation of human attributes, e.g. head, pants and upper clothes, whereas several recent works \cite{2020MUST, 2020Deep, 2020Controllable} treat the clothes as a type of texture style and try to encode it. Especially, thanks to the high flexibility and strong generalization ability to other clothes styles \cite{2020Controllable, 2019A}, the \emph{de facto standard of} Adaptive Instance Normalization (AdaIN) and its variants \cite{2017Arbitrary} have witnessed remarkable improvements in extracting appearance features \cite{2020MUST, 2020Controllable}. While obtaining acceptable performance, they have obvious drawbacks: they can often only perform coarse-grained feature extraction, and cannot restore fine-grained details well, which easily causes blurry clothes and the shortage of identity information. In the field of video-based generation, existing methods formulate this problem as a generation process of image sequence and produce results in a frame-by-frame manner. For example, resorting to a set of sparse trajectories, studies~\cite{2019Animating} and~\cite{2020First} leverage zeroth-order and first-order Taylor expansions such that the complex transformations are approximated. However, they rarely consider the video spatio-temporal coherence and thus affects the consistency of movements. Albeit the relatively smooth videos, a sequential generator proposed by Wang et al.~\cite{2019Few} is solely applied in tasks of single-person pose transfer.

Such limitations and drawbacks drive our exploration of clothes style transfer for person video generation to improve the quality and consistency of the generated video. In this paper, we propose a novel framework for person video generation towards using clothes style transfer. It consists of four components: \emph{Pose Landmark Encoder, Characteristic Identity Encoder, Clothes Style Encoder} and a \emph{Shared Decoder}, as illustrated in Figure~\ref{generator}. In specific, we formulate the task of feature learning as three disentangled sub-tasks to avoid the learning conflicts existing in the single-column structures. The disentangled multi-branch scheme is inspired by the natural insight that different salient areas correspond to varying degrees of responses to multiple tasks \cite{2020Revisiting}. Furthermore, to guarantee the spatio-temporal consistency in generated videos and implicitly capture the local frames dependencies, we delicately devise a variant of discriminator, named as \emph{inner-frame discriminator}, which takes the cross-frame difference as input. Besides, it is worth noting that background/scenarios replacement has also been a new trend for video person generation since it has higher values of privacy and entertainment~\cite{2020Real}. Unfortunately, existing approaches~\cite{2020MUST, 2020Controllable} are scenario-agnostic. Motivated by this observation, we design a strategy to expand our framework's capability of being scenario-aware and improve the flexibility of video generation. To our knowledge, this method provides the first investigation into spatio-temporal consistency and scenario awareness simultaneously for person video generation with clothes style transfer. Extensive experiments are conducted on TEDXPeople benchmark \cite{2021StylePeople} and demonstrate that our proposed method outperforms existing popular approaches in terms of image quality, similarity to realistic images and video smoothness, while embracing the merit of scenario awareness for matting.

\vspace{-0.35cm}
\section{Methodology}

\vspace{-0.3cm}
\subsection{Disentangled Multi-branch Encoders}
\vspace{-0.2cm}
To prevent the model from suffering from learning conflicts often occurring in single-column structure and capture fine-grained cues for the downstream decoder, we disentangle different spatial information from the main input and construct a multi-branch structure, which takes the pose landmark images $I_{p}^t$, characteristic identity images $I_{i}^t$ and clothes style images $I_c$ as input. Note the pose landmark images $I_{p}^t$ are obtained using a commonly-used human pose estimation method proposed by~\cite{openpose}. Following previous works~\cite{2020MUST, 2020Deep, 2020Controllable}, we build our \textbf{{pose landmark encoder}} upon a three-layer down-sampling network to encode the pose frame sequence. The {\textbf{characteristic identity encoder}} aims to boost the identity information and get rid of possible distortion, especially in the face region. Specifically, the identity images $I_{i}^t$ are disentangled by leveraging a human segmentation method in~\cite{2018Encoder}. Unlike the pose landmark encoder with demands of extracting more abundant and sophisticated features, the identity encoder plays a role of providing weak cues for identity information, and thus we narrow the network width to half of the pose encoder while reducing the number of parameters. For the {\textbf{clothes style encoder}}, the same segmentor~\cite{2018Encoder} is also employed to generate the clothes input images $I_{c}$. Besides, during the training phrase, a stochastic regularizer of geometric transformation is adopted to strengthen the model's generalization ability to arbitrary clothes input. In addition, to further increase model's capacity of learning representations, the combination of spatial and channel attention mechanisms~\cite{2018CBAM} is involved in each layer of encoders to adaptively emphasize the informative filters or regions by softly recalibrating corresponding units.


\vspace{-0.3cm}
\subsection{Shared Decoder with Scenario Awareness}
\vspace{-0.2cm}
After flowing through  multi-branch encoders in parallel, the disentangled sources are converted into the fine-grained feature representations. Then, these features are integrated and further processed in the high-level shared decoder. It is worth noting that skip connections are used here to fuse multi-level features while avoiding the issue of gradient vanishing \cite{2016Deep}. To let the framework possess higher flexibility of arbitrarily changing scenarios, we design an auxiliary output branch to predict the scenario mask $\alpha$. The final composition result can be formulated as $I_{composite}=\alpha I_{out}+(1 - \alpha)I_{bg}$, where $I_{out}$ and $\alpha$ denote the generated frame and the alpha matte for new scenarios, respectively, whereas $I_{bg}$ represents the arbitrary scenario image.





\vspace{-0.3cm}
\subsection{Inner-frame Discriminator}
\vspace{-0.2cm}
In general,  two kinds of discriminators ($D_{P}$ and $D_{I}$) are prevailing in current algorithms for pose-guided person image generation, such as~\cite{2020Controllable, 2019Progressive, 2018Unsupervised}. In detail,  $D_{P}$ is leveraged to guarantee the consistency between the pose of generated image $\hat{I}_{s}^t$, and the target pose $I_{P}^t$, while $D_{I}$ concentrates on the quality and reality of the generated images. To make full use of merits of both discriminators, both of them are adopted in our framework.



\begin{figure}[htbp]

\centering
\includegraphics[width=8.5cm]{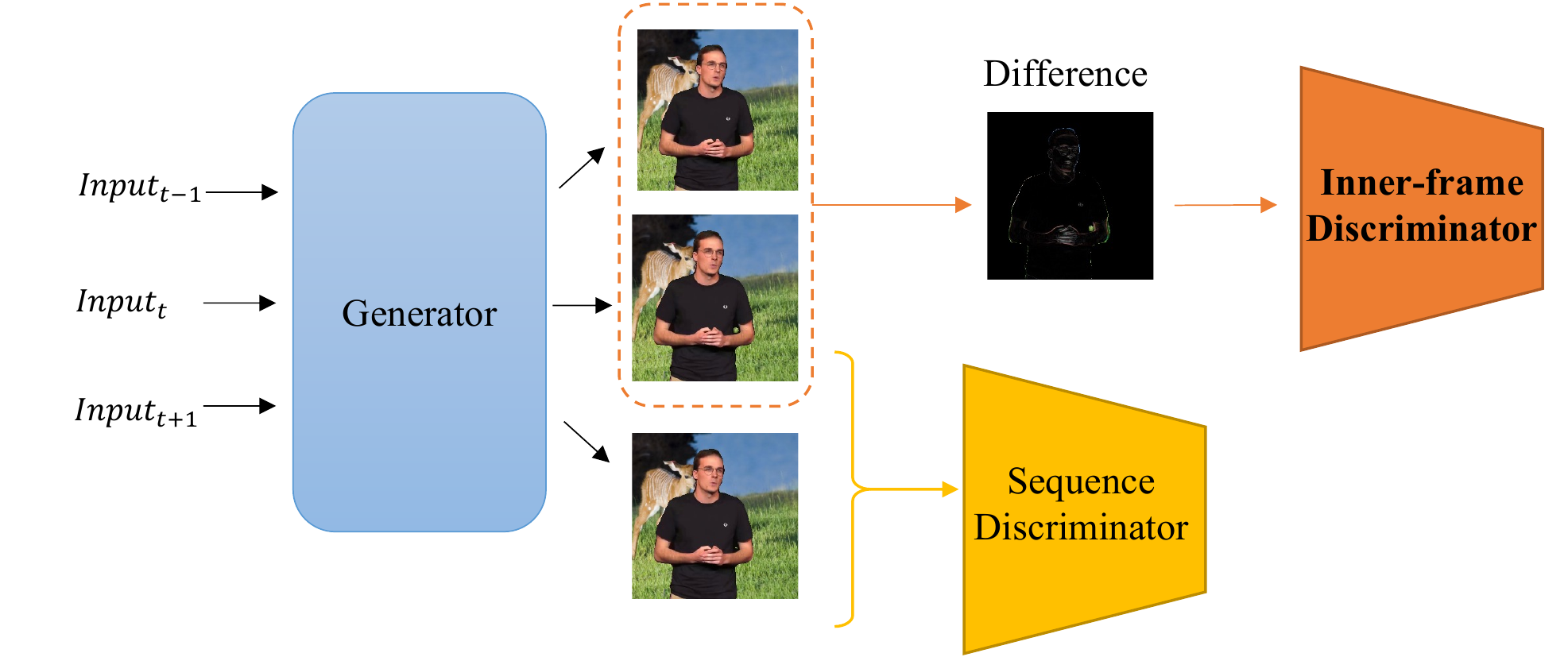}
\caption{The overview of the proposed inner-frame discriminator for spatio-temporal consistency in generated videos, which is accompanied by the sequence discriminator.}
\label{discriminator}
\end{figure}

\begin{figure*}[htbp]
\centering
\includegraphics[width=15cm, height=7cm]{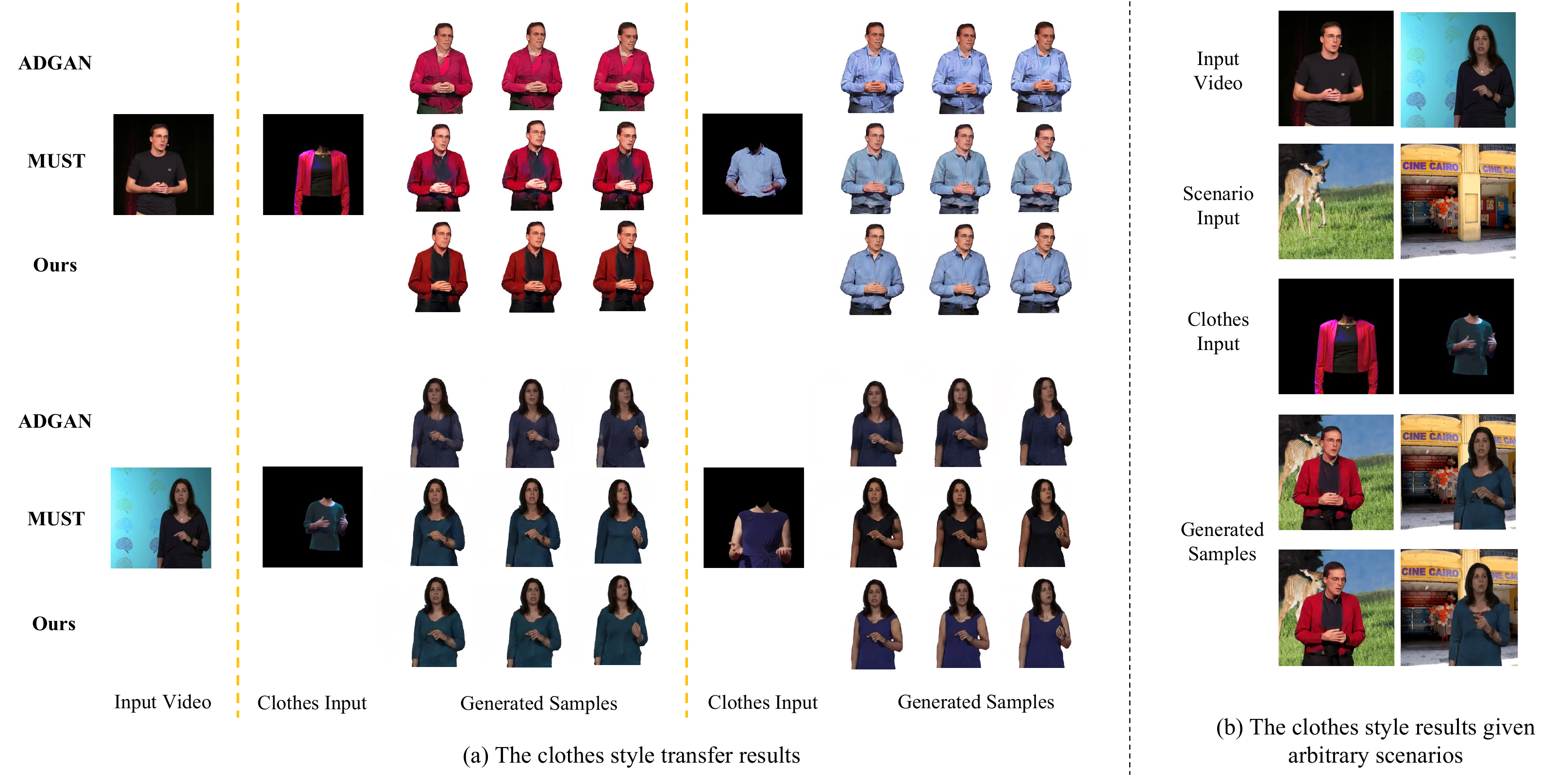}
\caption{The visualization examples generated by our method on TEDXPeople dataset.}
\label{comparison}
\end{figure*}

As described in Section~\ref{sec:intro}, spatio-temporal consistency of generated videos is short of being considered in existing approaches. To alleviate this issues, we devise a \textit{\textbf{inner-frame discriminator $D_{if}$}} in parallel with $D_{P}$ and $D_{I}$, which takes the cross-frame difference as the input and contributes to better video smoothness. In other words, it operates as an strong constraint imposed on adjacent frames to suppress suddenly drastic changes. Formally, given two adjacent frames $\hat{I}_{s}^t$ and $\hat{I}_{s}^{t+1}$ in the same video, the inner-frame discriminator can be denoted as: $D_{if}(\hat{I}_{s}^{t+1} - \hat{I}_{s}^{t})$. Besides, the sequence discriminator $D_t$ in \cite{2020Deep, 2019Few} is also incorporated into our method to enhance the visual quality of the generated videos since it enables the framework to implicitly consider the inner-relationships among frames. Figure~\ref{discriminator} delineates the designed scheme of these two discriminators.


\vspace{-0.3cm}
\subsection{Objective Functions}
\vspace{-0.2cm}
The entire training objective is to supervise the optimization of our proposed framework for the sake of high-quality person videos generation conforming to the input clothes style as well as the identity. Additionally, the posture of the generated person should be compatible with that of input pose images. The elaborate loss function of the whole network is formulated as: 
\begin{equation}
L = \lambda_{adv}L_{adv} + \lambda_{r}L_{r} + \lambda_{s}L_{s} + \lambda_{v}L_{v} + \lambda_{l}L_{l}
\end{equation}
where $\lambda_{adv}, \lambda_{r}, \lambda_{s}, \lambda_{v}, \lambda_{l}$ are the weights which are used to control the influence of different loss items. $L_{adv}$ indicates the adversarial loss implemented by the LSGAN \cite{2017Least} loss. Following \cite{2016Perceptual}, the reconstruction loss $L_r$ provides the supervision to impel the generated frames to be close to the ground truths following \cite{2016Perceptual} and \cite{2014Very} while style loss $L_s$ aims to maintain style coherence. More importantly, the video loss $L_v$ is achieved by discriminators $D_t$ and $D_{if}$ illustrated in Section~\ref{discriminator} to facilitate the video quality, especially for spatio-temporal coherence. Finally, it is worth noting that the truthfulness degree and identity preservation of generated images heavily rely on the synthesis quality of a family of key positions, such as head and hands parts. Therefore, to aid in improving comprehensive generation quality, we introduce a type of local characteristic discriminator which takes the head/hands segmentation results as input and computes the local characteristic loss term $L_{l}$.

\vspace{-0.3cm}
\section{Experiments}
\vspace{-0.2cm}
\subsection{Dataset and Implementation Details}
\vspace{-0.2cm}
We evaluate our method on the TEDXPeople dataset provided in \cite{2021StylePeople}. This dataset contains 48,188 videos of TED and TED-X talks. We randomly select 5,000 videos from the entire dataset with a resolution of 256×256. 85$\%$ of the dataset is used as training data. The rest is used to validate the performance of clothes style transfer during the inference phase. Besides, images of ImageNet dataset \cite{ILSVRC15} are used as random scenario inputs during training process. The weights for the loss terms are set to $\lambda_{adv} = 1, \lambda_{r} = 5, \lambda_{s} = 50, \lambda_{v} = 1.5, \lambda_{l} = 2$. We adopt Adam optimizer \cite{2014Adam} with the momentum set to 0.5 to train our model for around 100k iterations.


\vspace{-0.5cm}
\subsection{Comparison with SOTA Models}
\vspace{-0.2cm}
In our experiments, Structured similarity (SSIM) \cite{2004Image} and Peak Signal-to-Noise Ratio (PSNR) are employed to measure the similarity between the generated images and the ground truths, whereas Fréchet Inception Distance (FID) \cite{2017GANs} is used to assess the realism and consistency of generated images \cite{2020MUST, 2020Deep}. We evaluate the spatio-temporal consistency of the generated videos via the Fréchet Video Distance (FVD) \cite{2018FVD}. We compare our method with two existing popular approaches MUST \cite{2020MUST} and ADGAN\cite{2020Controllable}. The quantitative results are demonstrated in Table~\ref{tab:comparison} while the visualization examples are also provided in Figure 3(a). Both quantitative results and visualization examples demonstrate the superiority of our proposed framework. Meanwhile, we further validate the scenario awareness of our framework and several example results are given in Figure 3(b). As it can be observed, the identity of the generated videos remain the same as the source persons, and our approach can manipulate clothes styles and background scenarios flexibly. Video demos are available at \url{https://github.com/XSimba123/demos-of-csf-sa/}


\begin{table}[htbp]
\centering
\caption{Comparison with SOTA models on TEDXPeople. The proposed method consistently surpasses the existing approaches on all metrics. }
\begin{tabular}{c|c|c|c|c}
\hline
Method & SSIM $\uparrow$ & PSNR $\uparrow$ & FID $\downarrow$ & FVD $\downarrow$ \\ \hline
ADGAN \cite{2020Controllable} & 0.787 & 19.0 & 19.1 & 0.732\\ 
MUST \cite{2020MUST} & 0.794 & 19.3 & 11.7 & 0.804   \\ 
Our approach & \textbf{0.841} & \textbf{23.9} & \textbf{11.3} & \textbf{0.261} \\ \hline
\end{tabular}
\label{tab:comparison}
\end{table}



\vspace{-0.4cm}
\subsection{Ablation Study}
\vspace{-0.2cm}
To better understand each component in our model, we conduct ablation studies on TEDXPeople benchmark (see Table~\ref{tab:ablation}), which demonstrates that the multiple encoders structure (I) and attention module (II) have a great contribution to visual quality. The FVD metric shows that two spatio-temporal constraints (III) brings a significant improvement to video consistency. And local characteristic discriminator (IV) also gives a boost to the similarity metrics.
\begin{table}[htbp]
\centering

\caption{Ablation on disentangled multi-branch encoders (I), attention mechanism (II), spatio-temporal constraints (III) and local characteristic loss (IV).}
\begin{tabular}{c c c c|c|c|c|c}
\hline

I & II & III & IV & SSIM $\uparrow$ & PSNR $\uparrow$ & FID $\downarrow$ & FVD $\downarrow$ \\ \hline
 &  &  &  & 0.801 & 20.5 & 16.7 & 0.643\\ 
\Checkmark &  &  &  & 0.804 & 21.5 & 14.6 & 0.498\\ 
\Checkmark & \Checkmark &  &  & 0.818 & 21.8 & 12.8 & 0.427\\ 
\Checkmark & \Checkmark & \Checkmark &  & 0.826 & 22.3 & 12.9 & 0.282\\ 
\Checkmark & \Checkmark & \Checkmark & \Checkmark & \textbf{0.841} & \textbf{23.9} & \textbf{11.3} & \textbf{0.261} \\ \hline
\end{tabular}
\label{tab:ablation}
\end{table}

\vspace{-0.8cm}
\section{Conclusion}
\vspace{-0.3cm}
In this work, we propose an approach to tackle scenario-aware clothes style transfer for person videos, which is shown to outperform baseline and recent approaches. This is because our model uses a multi-encoder framework to handle different types of features. And an inner-frame discriminator is proposed to greatly enhance the video temporal coherence. Moreover, we design a framework to support scenario-aware video generation. Experimental results on TEDXPeople dataset confirmed the effectiveness and robustness of our proposed framework.

\newpage
\vfill\pagebreak

\bibliographystyle{IEEEbib}
\bibliography{refs}

\end{document}